\documentclass[10pt,twocolumn,letterpaper]{article}

\usepackage{cvpr}              

\usepackage{makecell}

\definecolor{cvprblue}{rgb}{0.21,0.49,0.74}
\usepackage[pagebackref,breaklinks,colorlinks,allcolors=cvprblue]{hyperref}
\usepackage[accsupp]{axessibility}  

\usepackage{xcolor}

\def\paperID{57} 
\def\confName{CVPR}
\def\confYear{2026}

\title{MTLLFM: Multimodal-Temporal Laughter Localization: UR-FUNNY-Temporal and SMILE-Temporal Benchmarks with an Adaptive Multimodal Fusion Model}

\makeatletter
\def\@fnsymbol#1{\ensuremath{
  \ifcase#1\or *\or \S\or \ddagger\or \mathsection\or \mathparagraph\else\@ctrerr\fi}}
\makeatother

\author{
Eyal Hanania\thanks{Corresponding author: \texttt{Eyal.Hanania@wsc-sports.com}} \quad
Nadav Kirsch \quad
Daniel Arkushin \quad
Jonathan Benvenisti \\[3pt]
Amos Bercovich \quad
Elie Zemmour\footnotemark[2] \quad
Sahar Froim\thanks{Contributed equally as co-senior authors.} \\[6pt]
WSC-Sports
}

\begin{document} 
\maketitle

\begin{abstract}

Detecting laughter in video is essential for affective computing and narrative understanding, yet existing approaches treat it as coarse clip-level classification, failing to capture precise temporal boundaries of brief, transient laughter events. We address this gap with two complementary contributions.

First, we introduce UR-FUNNY-Temporal and SMILE-Temporal, fully annotated temporal laughter datasets extending two widely-used humor benchmarks. Our annotations cover over 11,053 videos (78.8 hours) and provide precise onset/offset boundaries for each laughter event, along with rich metadata distinguishing speaker vs. audience laughter, modality dominance (acoustic, visual, or both), and intensity levels.

Second, we propose a lightweight weakly-supervised framework for temporal laughter localization. Our architecture combines fixed HuBERT and MAE encoders with temporal softmax pooling and adaptive modality gating, learning fine-grained temporal grounding from clip-level labels without requiring frame-level annotations during training. Experiments across three datasets demonstrate that our approach substantially outperforms multimodal foundation models including Gemini 3 Flash, achieving 99\% F1 and 68.1\% localization precision on sports broadcast data. Ablations validate each architectural component. Furthermore, our precise temporal tags improve downstream laughter reasoning by 227\% on CIDEr, enabling GPT-3.5 to outperform GPT-4o. The code, UR-FUNNY-Temporal and SMILE-Temporal datasets are publicly available at \url{https://github.com/WSCSports/MTLLFM-temporal-laughter-localization}.

\end{abstract}

\section{Introduction}
\label{sec:intro}

Laughter is a fundamental social signal that conveys humor, agreement, sarcasm, and emotional engagement in human communication. Automatically detecting laughter in unconstrained video is therefore an important capability for affective computing \cite{cortinas2023toward, pei2024affective}, multimodal reasoning \cite{shou2025multimodal, kasu2025d}, and narrative understanding \cite{ranade2022computational}. Despite recent advances in multimodal perception and affect recognition \cite{ramaswamy2024multimodal,udahemuka2024multimodal}, most existing approaches treat laughter detection as a coarse clip-level classification problem. In real-world conversational and broadcast settings, however, laughter events are often brief, sporadic, and temporally localized within longer segments of neutral speech or background activity. This mismatch between annotation granularity and event duration introduces significant label noise and limits the ability of models to learn precise temporal representations.

In this work, we address the problem of \textbf{fine-grained temporal localization of laughter} in multimodal video streams. Unlike prior methods that aggregate information across entire clips \cite{liu2024funnynet,deng2025bcfnet,gao2024fef}, we focus on identifying the exact onset and offset of laughter events using complementary audio and visual cues. This formulation enables a range of downstream applications and provides a general framework for temporally grounding short-lived affective expressions in multimodal video.

To support this task, we propose a lightweight multimodal architecture that combines fixed audio and visual feature encoders with an adaptive temporal fusion mechanism. Importantly, the model is designed to learn fine-grained temporal localization \textbf{without requiring temporally accurate frame-level annotations during training}, following recent trends in weakly supervised temporal event localization \cite{vahdani2022deep,zhang2022twinnet,yao2023weakly}. In contrast to conventional cross-attention fusion mechanisms that jointly model all temporal interactions at high computational cost \cite{vaswani2017attention}, our approach employs a lightweight saliency-driven temporal aggregation that explicitly focuses on short-lived affective peaks. This design enables precise temporal grounding while maintaining computational efficiency suitable for large-scale continuous video analysis. Although focused on laughter, the framework naturally extends to other transient affective signals such as excitement bursts or subtle emotional cues across diverse video domains.

In addition, we introduce temporally refined annotations for laughter events derived from widely used multimodal humor and affect datasets - UR-FUNNY and SMILE \cite{hasan2019urfunny,hyun2024smile}. Unlike existing annotations that label entire clips as humorous or non-humorous, our dataset provides \textbf{accurate temporal boundaries} for laughter occurrences. Our annotations further distinguish between \textbf{speaker laughter and audience laughter}, and identify whether the dominant perceptual cue is \textbf{audio, visual, or jointly multimodal}. This enriched supervision enables more rigorous evaluation of temporal localization methods and facilitates new research directions in fine-grained multimodal affect understanding.

Experimental results demonstrate that the proposed approach significantly improves temporal localization accuracy compared to strong multimodal baselines and recent multimodal foundation models. Our findings highlight the importance of task-specific temporal modeling and suggest that lightweight architectures can outperform general-purpose reasoning systems when precise temporal grounding is required.

\noindent\textbf{Contributions.}
\begin{itemize}
\item We propose a \textbf{lightweight multimodal architecture for weakly supervised temporal localization of laughter} that enables precise onset detection even when temporally accurate annotations are unavailable during training.
\item We introduce a \textbf{temporally refined laughter datasets} extending existing multimodal humor benchmarks, providing accurate onset and offset boundaries for laughter events.
\item Our annotations further distinguish between speaker laughter and audience laughter, and explicitly label the dominant perceptual cue (audio, visual, or joint multimodal), enabling new research directions in multimodal affect grounding and modality-aware reasoning.
\end{itemize}

\section{Related Work}

\subsection{Laughter and Humor Recognition}

Automatic laughter and humor recognition has been explored using both acoustic and visual cues. Early audiovisual laughter detection works demonstrated that combining facial and acoustic information improves over single-modality approaches \cite{petridis2008fusion,escalera2009multimodal}. More recent multimodal humor research has shifted toward clip-level understanding in conversational settings, with datasets such as UR-FUNNY enabling learning from text, audio, and visual streams \cite{hasan2019urfunny}. SMILE further expands this space by introducing video-language reasoning about laughter, but still operates at the clip level rather than temporally precise event localization \cite{hyun2024smile}.

\subsection{Multimodal Affective Video Understanding}

Multimodal affect recognition has benefited from advances in deep representation learning, benchmark construction, and increasingly expressive fusion mechanisms \cite{poria2017review,kollias2022abaw}. Representative approaches include tensor-based fusion \cite{zadeh2017tensor}, cross-modal transformers \cite{tsai2019multimodal}, modality-invariant and modality-specific representation learning \cite{hazarika2020misa}, and adaptation modules for large pretrained transformers \cite{rahman2020integrating}. While these methods have advanced multimodal affect and sentiment understanding, they are typically optimized for clip- or utterance-level prediction rather than isolating short-lived affective events within longer neutral context.

\subsection{Temporal Event Localization}

Temporal localization has been extensively studied in action recognition and video understanding, where the goal is to identify precise event boundaries within long untrimmed videos. Weakly supervised formulations, such as UntrimmedNets, have shown that latent selection mechanisms can recover meaningful temporal structure without dense boundary labels \cite{wang2017untrimmednets}. However, analogous fine-grained temporal modeling remains relatively underexplored in affective computing, especially for transient multimodal behaviors such as laughter.



\section{Method}
\label{sec:method}

\subsection{Problem Formulation}
\label{sec:problem_formulation}

We formulate laughter detection as a weakly-supervised temporal localization problem. Given a video segment of duration $T$ seconds, let $\mathbf{a} = \{a_1, \ldots, a_{T_a}\}$ and $\mathbf{v} = \{v_1, \ldots, v_{T_v}\}$ denote the audio and visual feature sequences with $T_a$ and $T_v$ timesteps respectively, where $a_t \in \mathbb{R}^{d_a}$ and $v_t \in \mathbb{R}^{d_v}$. Our model learns a mapping:
\begin{equation}
f: (\mathbf{a}, \mathbf{v}) \;\rightarrow\; (\hat{y},\; \boldsymbol{\alpha}^{a},\; \boldsymbol{\alpha}^{v},\; w_a,\; w_v)
\end{equation}
where $\hat{y} \in \{0, 1\}$ is the predicted binary label, $\boldsymbol{\alpha}^{a} \in \mathbb{R}^{d_a}$ and $\boldsymbol{\alpha}^{v} \in \mathbb{R}^{d_v}$ are temporal attention distributions over each modality, and $w_a, w_v \in [0, 1]$ are learned modality weights satisfying $w_a + w_v = 1$.

During training, we assume access only to clip-level binary labels indicating whether laughter occurs anywhere within the segment, without frame-level annotations of exact onset and offset times. At inference, the learned attention distributions provide implicit temporal grounding, enabling post-hoc localization by identifying peak attention timesteps. The central challenge lies in designing an architecture capable of isolating brief, transient laughter events from surrounding neutral context using only weak clip-level supervision.

\subsection{Architecture Overview}
\label{sec:architecture}

\begin{figure}[t]
\centering
\includegraphics[width=\columnwidth]{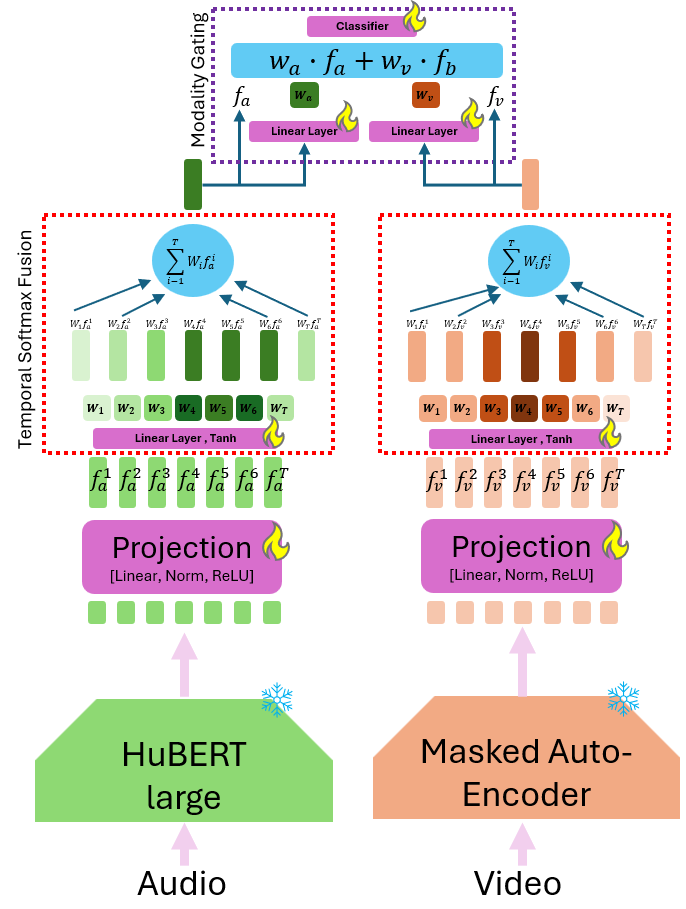}
\caption{Architecture overview. Frozen HuBERT and MAE encoders extract audio and visual features, which are projected to a shared hidden space $\mathbb{R}^d$. Temporal Softmax Fusion independently pools each modality by learning attention distributions $\boldsymbol{\alpha}^a, \boldsymbol{\alpha}^v$ over timesteps. Adaptive Modality Gating produces complementary weights $w_a, w_v$ to fuse the pooled representations $\mathbf{f}^a, \mathbf{f}^v$. A final classifier outputs the binary laughter prediction $\hat{y}$.}
\label{fig:architecture}
\end{figure}

Our architecture consists of four main components illustrated in Figure~\ref{fig:architecture}: fixed feature encoders, modality-specific projection and temporal pooling, adaptive gating for multimodal fusion, and a classification head.

\textbf{Feature Extraction.} We employ frozen pre-trained encoders to extract complementary multimodal representations. HuBERT-Large~\cite{hsu2021hubert} processes the audio stream at approximately 50Hz, producing a sequence of 1024-dimensional features. In parallel, a Masked Auto-Encoder (MAE)~\cite{he2022mae} extracts 768-dimensional visual features at 10fps. Using fixed encoders significantly reduces computational costs and training complexity while leveraging robust representations learned from large-scale pretraining. This encoders combination aligns with prior affective multimodal work~\cite{cheng2024emotionllama}.


\textbf{Temporal Fusion.} Each modality is processed independently through a projection layer followed by a temporal softmax pooling module. The projection layer (Linear + LayerNorm + ReLU) maps features from each modality to a shared 1024-dimensional hidden space. The temporal softmax pooling mechanism then computes a learned attention distribution over timesteps, enabling the model to focus on salient emotional peaks while suppressing neutral background context. This produces fixed-dimensional representations $\mathbf{f}^a$ and $\mathbf{f}^v$ for audio and visual streams respectively.

\textbf{Adaptive Modality Gating.} To handle cases where laughter manifests primarily through one modality, we employ a learnable gating mechanism that dynamically weights the contribution of each modality. Two independent linear layers compute gate logits from $\mathbf{f}^a$ and $\mathbf{f}^v$, which are then normalized via softmax to produce complementary modality weights $w_v$ and $w_a$. The fused representation $\mathbf{f}_{\text{fused}} = w_a \cdot \mathbf{f}^a + w_v \cdot \mathbf{f}^v$ allows the model to prioritize the more reliable modality for each instance, as detailed in Section~\ref{sec:gating}.

\textbf{Classification.} A final linear layer with dropout maps the fused representation to binary logits, trained with Focal Loss~\cite{lin2017focal} to handle class imbalance. The following subsections provide detailed formulations for each component.

\subsection{Temporal Softmax Pooling}
\label{sec:temporal_pooling}

A key challenge in weakly-supervised temporal localization is learning to focus on brief emotional events buried within longer neutral segments, when only clip-level labels are available. Standard temporal aggregation methods such as mean or max pooling treat all timesteps uniformly or select only the most extreme feature, failing to capture the nuanced temporal structure of transient affective signals. We adapt the temporal softmax pooling mechanism from weakly-supervised action localization~\cite{nguyen2018weakly} to the affective domain, where events are significantly shorter and more subtle than typical action boundaries.

\textbf{Mechanism.} For each modality $m \in \{a, v\}$, let $\mathbf{F}^m = \{f_m^1, \ldots, f_{m}^{T_m}\}$ denote the projected feature sequence with $f_m^t \in \mathbb{R}^{d}$. We compute a scalar importance score for each timestep:
\begin{equation}
\label{eq:attn_score}
e_m^t = \tanh(\mathbf{w}^{\top} f_m^t + b)
\end{equation}
where $\mathbf{w} \in \mathbb{R}^{d}$ and $b \in \mathbb{R}$ are learned parameters. The $\tanh$ non-linearity constrains scores to $[-1, 1]$, preventing saturation during softmax normalization. These scores are normalized into a probability distribution over time:
\begin{equation}
\label{eq:attn_weight}
\alpha_m^t = \frac{\exp(e_m^t)}{\sum_{j=1}^{T_m} \exp(e_m^j)}
\end{equation}

The final pooled representation is obtained via weighted aggregation:
\begin{equation}
\label{eq:pooling}
\mathbf{f}_m = \sum_{t=1}^{T_m} \alpha_m^t \cdot f_m^t
\end{equation}

\textbf{Adaptation to Affective Signals.} While prior work applied this mechanism to action localization in untrimmed videos where events span seconds to minutes~\cite{nguyen2018weakly}, laughter events are significantly more transient (often sub-second bursts). We make two key adaptations: (1) we apply the pooling independently per modality before fusion, rather than on concatenated features, enabling modality-specific temporal grounding; and (2) we incorporate tanh gating before softmax to handle the higher temporal resolution and shorter event duration characteristic of affective signals.



\textbf{Properties.}. This mechanism operates in O(T) time complexity, computing a scalar score per timestep rather than modeling pairwise interactions as in self-attention (O(T²)), while providing interpretable temporal grounding through the learned attention distribution.

\subsection{Adaptive Modality Gating}
\label{sec:gating}

Laughter manifests differently across modalities depending on context. In some cases, laughter is predominantly audible (e.g., loud vocal laughter with minimal facial expression), while in others it is primarily visual (e.g., suppressed smiles or silent chuckling). Moreover, in sports broadcast settings, audio streams often contain deceptive commentary noise-high-energy speech that mimics emotional intensity-while visual streams may capture stoic on-screen subjects despite off-camera laughter. To handle these modality conflicts and asymmetries, we employ an adaptive gating mechanism~\cite{arevalo2017gated} that learns to dynamically weight the contribution of each modality based on its reliability for each instance.

\textbf{Mechanism.} Given the pooled representations $\mathbf{f}_a \in \mathbb{R}^{d}$ and $\mathbf{f}_v \in \mathbb{R}^{d}$ from the temporal pooling stage (Eq.~\ref{eq:pooling}), we compute modality-specific gate logits using independent linear projections:
\begin{equation}
\label{eq:gate_logits}
g_a = \mathbf{w}_a^{\top} \mathbf{f}_a, \quad g_v = \mathbf{w}_v^{\top} \mathbf{f}_v
\end{equation}
where $\mathbf{w}_a, \mathbf{w}_v \in \mathbb{R}^{d}$ are learnable parameters. These logits are normalized via softmax to produce complementary modality weights:
\begin{equation}
\label{eq:gate_weights}
[w_a, w_v] = \mathrm{softmax}([g_a, g_v])
\end{equation}
ensuring $w_a + w_v = 1$. The final fused representation is:
\begin{equation}
\label{eq:fusion}
\mathbf{f}_{\text{fused}} = w_a \cdot \mathbf{f}_a + w_v \cdot \mathbf{f}_v
\end{equation}



\subsection{Inference and Localization}
\label{sec:inference}

At inference time, our model produces three outputs for each video segment: a predicted label $\hat{y}$, modality-specific attention distributions $\boldsymbol{\alpha}^{a}$ and $\boldsymbol{\alpha}^{v}$, and modality weights $w_a$ and $w_v$. For segments classified as containing laughter ($\hat{y} = 1$), we exploit these attention patterns to perform post-hoc temporal localization.

\textbf{Localization Procedure.} Given attention vectors $\boldsymbol{\alpha}_{a} \in \mathbb{R}^{d_a}$ and $\boldsymbol{\alpha}_{v} \in \mathbb{R}^{d_v}$, we first apply temperature sharpening to accentuate peaks:
\begin{equation}
\label{eq:sharpening}
\tilde{\alpha}_m^t = \frac{(\alpha_m^t)^{1/\tau}}{\sum_{j=1}^{T_m} (\alpha_m^j)^{1/\tau}}
\end{equation}
where $\tau \in (0, 1]$ controls the sharpening degree (lower $\tau$ produces sharper distributions). We then align both modalities to a common resolution of $N$ temporal bins via max-pooling (audio) and linear interpolation (visual). The modality-weighted combined signal is:
\begin{equation}
\label{eq:combined_attention}
\beta_n = w_a \cdot \tilde{\alpha}_a^{(n)} + w_v \cdot \tilde{\alpha}_v^{(n)}, \quad n = 1, \ldots, N
\end{equation}
where $\tilde{\alpha}_m^{(n)}$ denotes the aligned attention value for modality $m$ at bin $n$.

We identify the peak bin $n^* = \arg\max_n \beta_n$ and expand the predicted interval by including adjacent bins whose combined attention exceeds the mean $\bar{\beta} = \frac{1}{N}\sum_{n=1}^{N} \beta_n$. The bin range is then mapped to continuous timestamps within the segment. In our experiments, we use $\tau = 0.5$ and $N = 10$ bins for 5-second segments.

\section{Temporal Laughter Annotations}
\label{sec:dataset}


A significant contribution of this work is the creation of fine-grained temporal laughter annotations for the complete UR-FUNNY~\cite{hasan2019urfunny} and SMILE~\cite{hyun2024smile} datasets. While these datasets provide valuable resources for humor recognition research, their original annotations are limited to clip-level binary labels indicating whether humor is present. This coarse labeling scheme is insufficient for training and evaluating models that require precise temporal localization of laughter events. Moreover, existing laughter datasets lack the scope of ours: AudioSet~\cite{gemmeke2017audio} offers only weak clip-level labels; Switchboard~\cite{godfrey1992switchboard} is audio-only conversational data; and recent stand-up comedy datasets~\cite{kuznetsova2024multimodal,barriere2025standup4ai} use automatic annotations without modality-dominance or speaker-type metadata. None cover a multimodal broadcast setting with manually verified onset/offset boundaries. We address these limitations by manually annotating the exact onset and offset times of laughter occurrences for \textbf{all 10,166 videos in UR-FUNNY} and \textbf{all 887 videos in SMILE}, along with additional metadata characterizing each event. We refer to our temporally-annotated versions as \textbf{UR-FUNNY-Temporal} and \textbf{SMILE-Temporal}.

\subsection{Annotation Protocol}

Laughter events are manually annotated using frame-accurate audiovisual review. Temporal boundaries are defined from the first perceptible acoustic or visual cue to the complete cessation of the laughter signal, ensuring full coverage of each affective episode. 

For each event, annotators additionally label (i) speaker vs.\ audience origin, (ii) modality dominance (acoustic, visual, or both), and (iii) intensity (chuckle vs.\ sustained laughter). All annotations are performed by trained annotators with consistency verification.

\subsection{Dataset Statistics}

Table~\ref{tab:dataset_stats} summarizes the temporal laughter annotations for both datasets. We provide annotations for all 10,166 videos in UR-FUNNY-V2 (spanning 72.9 hours) and all 887 videos in SMILE (spanning 5.9 hours). Of these, 2,369 UR-FUNNY videos and 589 SMILE videos contain laughter, yielding 3,385 and 1,560 annotated laughter events respectively. The mean laughter duration is 1.70 seconds for UR-FUNNY and 2.16 seconds for SMILE, reflecting the transient nature of these affective signals.

The annotations reveal substantial asymmetry in modality dominance. In UR-FUNNY, 79.4\% of laughter events are primarily acoustic, 6.1\% are primarily visual, and 14.5\% exhibit strong multimodal signals. SMILE shows even stronger acoustic dominance (92.5\%), with only 1.5\% being primarily visual. This distribution validates the importance of adaptive gating: while audio is often more reliable, a non-trivial fraction of events require visual evidence or multimodal fusion.

Speaker vs. audience analysis shows that 81.6\% of UR-FUNNY laughter originates from the audience, compared to 93.7\% in SMILE. This reflects the conversational nature of the datasets, where audience reactions often provide stronger affective cues than the speaker's own expressions. The intensity distribution varies across datasets: UR-FUNNY contains 59.4\% chuckles and 40.6\% full laughter, while SMILE is dominated by more pronounced laughter events (71.3\%).

These annotations enable rigorous evaluation of temporal localization methods and support future research directions in modality-aware affect understanding, speaker-audience interaction dynamics, and fine-grained intensity estimation.

\subsection{Sports Broadcast Dataset}

To validate our approach on the target domain of sports broadcasting, we additionally curate a proprietary dataset, \textbf{SportsPress}, consisting of 3,787 5-second video segments extracted from sports press conferences and studio shows. The dataset contains 883 segments with laughter and 2,904 negative samples, providing a balanced evaluation set for the challenges described in Section~\ref{sec:intro}: deceptive commentary noise from high-energy sports commentary, off-camera laughter triggers while on-screen subjects remain stoic, and rapid modality conflicts typical of broadcast environments. We use this dataset for detailed ablation studies and baseline comparisons to demonstrate the effectiveness of our approach in real-world production settings. While we do not release SportsPress publicly due to broadcast rights constraints, our temporal annotation methodology and the released UR-FUNNY-Temporal and SMILE-Temporal annotations enable the research community to reproduce and extend our approach to other domains.

\subsection{Data Availability}

The UR-FUNNY-Temporal and SMILE-Temporal datasets, containing temporal annotations and metadata as CSV files, are publicly available for non-commercial research purposes at \url{https://github.com/WSCSports/MTLLFM-temporal-laughter-localization}. The underlying video content is part of the original UR-FUNNY and SMILE datasets. While our model uses only clip-level labels during training, we release complete temporal annotations for all splits to enable future work on fully-supervised and semi-supervised temporal localization of laughter.

\begin{table}[t]
\centering
\setlength{\tabcolsep}{8pt}
\renewcommand{\arraystretch}{1.15}
\caption{Statistics of temporal laughter annotations for UR-FUNNY-Temporal and SMILE-Temporal.}
\label{tab:dataset_stats}
\footnotesize
\resizebox{\columnwidth}{!}{%
\begin{tabular}{lrr}
\toprule
& \makecell{\textbf{UR-FUNNY}\\\textbf{Temporal}} & \makecell{\textbf{SMILE}\\\textbf{Temporal}} \\
\midrule
Total videos            & 10,166 & 887 \\
Total hours             & 72.9   & 5.9 \\
Videos with laughter    & 2,369  & 589 \\
Total laughter events   & 3,385  & 1,560 \\
Mean duration (s)       & 1.70   & 2.16 \\
Duration std dev (s)    & 1.45   & 1.51 \\
\midrule
\multicolumn{3}{l}{\textit{Speaker vs.\ Audience}} \\[2pt]
\quad Speaker  &   624  &    98 \\
\quad Audience & 2,761  & 1,462 \\
\midrule
\multicolumn{3}{l}{\textit{Modality Dominance}} \\[2pt]
\quad Acoustic & 2,686 & 1,443 \\
\quad Visual   &   207  &    24 \\
\quad Both     &   492  &    93 \\
\midrule
\multicolumn{3}{l}{\textit{Intensity Levels}} \\[2pt]
\quad Chuckle  & 2,012  &   447 \\
\quad Laughter & 1,373  & 1,113 \\
\bottomrule
\end{tabular}%
}
\end{table}

\section{Experiments}
\label{sec:experiments}

\subsection{Experimental Setup}

We evaluate our approach on three datasets: SportsPress, UR-FUNNY-Temporal, and SMILE-Temporal, using the original train/validation/test splits. Training is performed on 5-second video segments with binary clip-level labels (laughter present or absent). At inference, we evaluate classification performance on all test samples and temporal localization on positive samples with ground truth boundaries.

\subsection{Implementation Details}

\textbf{Feature Extraction.} We extract audio features using HuBERT~\cite{hsu2021hubert}, producing 1024-dimensional embeddings at approximately 50Hz (250 frames per 5-second segment). Visual features are extracted using MAE~\cite{he2022mae}, producing 768-dimensional embeddings at 10fps (50 frames per segment). All features are precomputed and cached.

\textbf{Training Procedure.} We train using Adam with learning rate $10^{-4}$ and batch size 32. Projection and fusion layers use hidden dimension 1024 with dropout 0.5. The model is trained for up to 50 epochs with early stopping based on validation loss. We use Focal Loss~\cite{lin2017focal} with class-specific weights reflecting the filtered data distribution.

\textbf{Localization Post-Processing.} At inference, we extract attention distributions from both modalities and align them to 50 temporal bins via max-pooling (audio) while visual attention is already at this resolution (10fps $\times$ 5 seconds). We combine the aligned attention using learned modality weights as described in Section~\ref{sec:inference}. The predicted interval is obtained via peak expansion: we identify the bin with maximum attention and expand left/right while attention exceeds the mean value, yielding temporal resolution of 0.1 seconds per bin.

\subsection{Ablation Variants}

To isolate each component's contribution, we train the following variants on SportsPress with identical hyperparameters (Table~\ref{tab:wsc_ablation}): \textbf{(1) Pooling} - mean, max, and self-attention~\cite{vaswani2017attention} pooling, each replacing our softmax pooling entirely; \textbf{(2) Temporal attention} - softmax pooling without tanh gating; \textbf{(3) Fusion strategy} - concatenation (no gating), sigmoid gating, and cross-attention fusion between modalities; \textbf{(4) No projection activation} - removing LayerNorm and ReLU from the projection layers; \textbf{(5) Single modality} - audio-only and vision-only variants.

\subsection{Multimodal Foundation Models}

We compare against three recent multimodal large language models: Gemini 3 Flash ~\cite{geminiteam2025gemini25}, Qwen2.5 Omni 7B (video and audio) ~\cite{qwen25omni2025}, and Qwen2 Audio 7B (audio-only) \cite{chu2024qwen2audio}. For each 5-second test segment, we prompt the models to identify whether laughter is present and, if so, to predict precise start and end timestamps in seconds. We evaluate these predictions using the same classification and localization metrics as our approach.

\subsection{Downstream Application: Video Laugh Reasoning}

To evaluate the practical utility of our temporal localization beyond detection metrics, we conduct an additional experiment on Video Laugh Reasoning~\cite{hyun2024smile}, which requires generating natural language explanations for why audiences laugh in video segments. We select 50 diverse clips from SMILE-Temporal and compare two large language models (GPT-3.5 and GPT-4o) under two input conditions: (1) \textit{Baseline}: raw transcript without temporal markers, and (2) \textit{Ours}: transcript augmented with \texttt{<LAUGHTER>} tokens inserted at precise onset timestamps predicted by our model. We evaluate generated explanations against ground-truth human annotations using standard language generation metrics: BLEU-1/4, METEOR, BERTScore, CIDEr, and ROUGE-L. Statistical significance is assessed via paired t-tests. Results and analysis are presented in Section~\ref{sec:downstream}.

\section{Experimental Results}
\label{sec:results}

We evaluate our approach on three datasets: SportsPress (our target domain), UR-FUNNY-Temporal, and SMILE-Temporal. We report both classification metrics (F1 score) and temporal localization metrics (Precision@IoU=0.5, Mean IoU). Localization metrics are computed only on positive samples with ground truth temporal boundaries.

\subsection{Comparison with Multimodal Foundation Models}

\begin{table*}[t]
\centering
\caption{Comparison across three datasets. Classification metrics computed on all test samples; localization metrics on positive samples only. Localization uses multi-GT IoU (maximum IoU against all ground-truth laughter events in each sample)}.
\label{tab:unified_final}
\footnotesize
\begin{tabular}{l|ccc|ccc|ccc}
\toprule
& \multicolumn{3}{c|}{\textbf{SportsPress}} & \multicolumn{3}{c|}{\textbf{UR-Funny}} & \multicolumn{3}{c}{\textbf{SMILE}} \\
\cmidrule(lr){2-4} \cmidrule(lr){5-7} \cmidrule(lr){8-10}
Method & Cls.F1 & Loc@.5 & IoU & Cls.F1 & Loc@.5 & IoU & Cls.F1 & Loc@.5 & IoU \\
\midrule
Qwen2 Audio 7B  & 0.982          & 0.028          & 0.059          & 0.466          & 0.137          & 0.172          & 0.335          & 0.138          & 0.204          \\
Qwen2.5 Omni 7B & \textbf{0.997} & 0.208          & 0.301          & 0.773          & 0.218          & 0.267          & 0.783          & 0.315          & 0.361          \\
Gemini 3 Flash  & 0.885          & 0.542          & 0.546          & 0.775          & 0.393          & 0.405          & 0.724          & \textbf{0.579} & \textbf{0.540} \\
\midrule
\textbf{MTLLFM (Ours)} & 0.990 & \textbf{0.681} & \textbf{0.580} & \textbf{0.849} & \textbf{0.497} & \textbf{0.466} & \textbf{0.803} & 0.567 & 0.511 \\
\bottomrule
\end{tabular}
\vspace{2mm}

\small
\textit{Note:} SportsPress ($N_\text{cls}$=260, $N_\text{loc}$=144); UR-Funny ($N_\text{cls}$=1,626, $N_\text{loc}$=789); SMILE ($N_\text{cls}$=921, $N_\text{loc}$=342).
\end{table*}

Table~\ref{tab:unified_final} compares our method against recent multimodal foundation models across all three datasets. While foundation models exhibit strong semantic reasoning capabilities, they face substantial challenges in precise temporal localization of brief affective events.

\textbf{SportsPress.} Our approach achieves 99.0\% classification F1 and 68.1\% Precision@IoU=0.5, substantially outperforming all baselines. Gemini 3 Flash achieves moderate localization (54.2\%) but lower classification accuracy (88.5\%). The Qwen models struggle severely with temporal grounding despite strong classification performance, with Qwen2.5 Omni achieving 99.7\% F1 but only 20.8\% localization precision. This demonstrates that semantic reasoning alone is insufficient for fine-grained temporal grounding in sports broadcast environments with deceptive commentary noise.

\textbf{UR-FUNNY-Temporal.} Our model achieves 84.9\% F1 and 49.7\% Precision@IoU=0.5, outperforming all foundation models on both classification and localization. The conversational nature of presentation-style humor presents greater challenges than sports broadcasts, reflected in lower absolute localization scores across all methods. Nevertheless, our learned temporal attention provides consistent advantages over global semantic reasoning.

\textbf{SMILE-Temporal.} Our approach achieves the highest classification F1 (80.3\%), while Gemini attains competitive localization performance (57.9\% vs. our 56.7\%). This dataset contains predominantly pronounced audience laughter in sitcom and talk show settings, where strong acoustic and visual cues may enable semantic models to infer temporal boundaries more reliably. However, our specialized architecture maintains superior classification while achieving comparable localization at significantly lower computational cost.

\textbf{Key Insights.} The results reveal a fundamental trade-off: foundation models offer semantic robustness but lack the fine-grained temporal precision required for brief affective events. Our task-specific architecture achieves superior temporal grounding through learned attention mechanisms, demonstrating that specialized temporal modeling outperforms general-purpose reasoning for sub-second event localization. The Qwen models' poor localization despite strong classification further validates that multimodal understanding does not automatically translate to precise temporal grounding.

\begin{table}[t]
\centering
\setlength{\tabcolsep}{6pt}
\renewcommand{\arraystretch}{1.1}
\caption{Ablation study on SportsPress (N$_\text{cls}$=261, N$_\text{loc}$=144).
$^{\S}$No attention weights; localization falls back to full-segment prediction.
$^{\P}$Equal modality weights ($w_a$=$w_v$=0.5) for localization.}
\label{tab:wsc_ablation}
\resizebox{\columnwidth}{!}{%
\begin{tabular}{lc|ccc}
\toprule
\textbf{Model Variant} & \textbf{F1}$\uparrow$ & \textbf{P@.5}$\uparrow$ & \textbf{IoU}$\uparrow$ & \textbf{Peak-GT}$\uparrow$ \\
\midrule
\multicolumn{5}{l}{\textit{Single Modality:}} \\[2pt]
\quad Audio Only                  & 0.979 & 0.604 & 0.552 & 0.764 \\
\quad Vision Only                 & 0.675 & 0.257 & 0.322 & 0.451 \\
\midrule
\multicolumn{5}{l}{\textit{Temporal Attention (softmax pooling kept):}} \\[2pt]
\quad w/o Tanh Gating             & 0.979 & 0.639 & 0.562 & 0.799 \\
\midrule
\multicolumn{5}{l}{\textit{Pooling (replaces softmax pooling):}} \\[2pt]
\quad Mean Pool$^{\S}$       & 0.968 & 0.160 & 0.347 & 0.590 \\
\quad Max Pool$^{\S}$        & 0.946 & 0.160 & 0.347 & 0.590 \\
\quad Self-Attention Pool         & 0.653 & 0.188 & 0.274 & 0.654 \\
\midrule
\multicolumn{5}{l}{\textit{Fusion Strategy (softmax pooling kept):}} \\[2pt]
\quad Concat (no gate)$^{\P}$ & 0.976 & 0.618 & 0.535 & 0.771 \\
\quad Sigmoid Gate                & 0.986 & 0.625 & 0.555 & 0.792 \\
\quad Cross-Attention Fusion      & 0.677 & 0.248 & 0.324 & 0.759 \\
\midrule
\multicolumn{5}{l}{\textit{Architecture:}} \\[2pt]
\quad No Proj.\ Activation        & 0.983 & 0.653 & 0.549 & 0.792 \\
\midrule
\textbf{Full Model (Ours)}        & \textbf{0.990} & \textbf{0.681} & \textbf{0.580} & \textbf{0.840} \\
\bottomrule
\end{tabular}%
}
\end{table}

\subsection{Ablation Analysis}
Table~\ref{tab:wsc_ablation} reports all ablation results.
Mean and max pooling (P@.5=.161) have no localization mechanism; our softmax pooling delivers a $4\times$ gain (68.1\%) with no classification cost.
Self-attention and cross-attention variants fall significantly short (P@.5=0.188 and 0.248, F1$\approx$0.65--0.68): general-purpose attention mechanisms require substantially more training data to learn the sparse, sub-second temporal structure of affective events under weak supervision.
Removing tanh gating costs 4.2 P@.5 points (63.9\%), confirming its role in preventing attention saturation.
Softmax gating outperforms both concatenation and sigmoid fusion by 3--7 P@.5 points, validating complementary modality weighting.
Finally, audio-only (60.4\%) far outperforms vision-only (25.7\%), consistent with the acoustic dominance in our annotations; the full model's additional +7.7-point gain confirms that adaptive gating captures complementary visual cues.

\begin{figure}[t]
\centering
\includegraphics[width=\columnwidth]{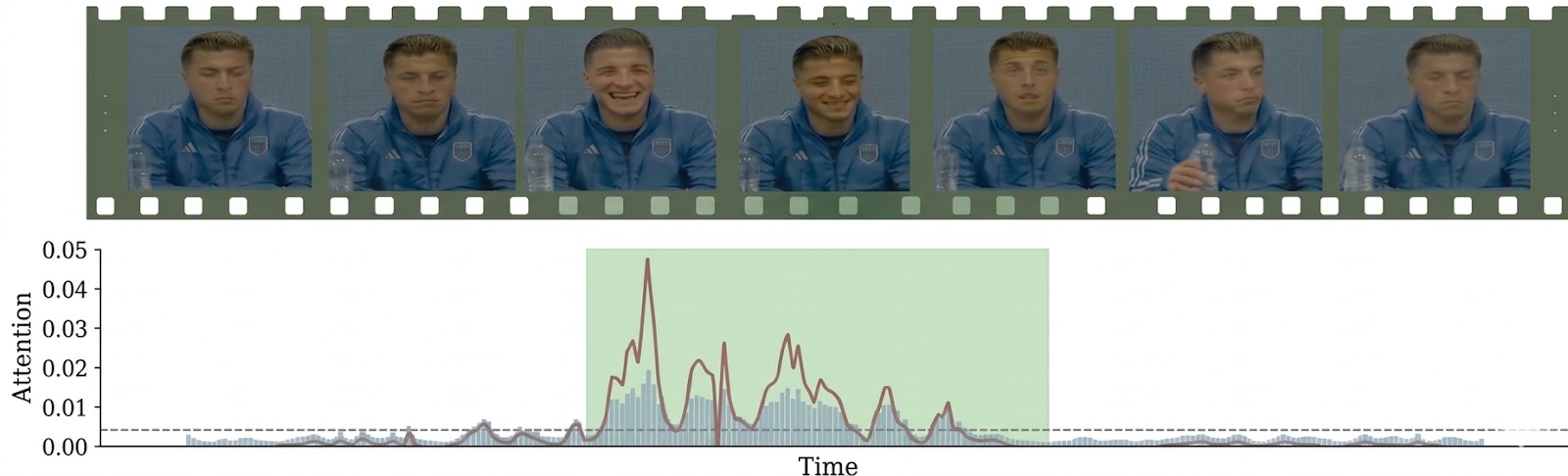}
\caption{Temporal localization via learned attention. The modality-weighted combined attention peaks precisely at the laughter event (green shaded region). Video frames show the speaker's expressions throughout the 5-second segment. The sharp peak demonstrates successful grounding from clip-level labels.}
\label{fig:attention_viz}
\end{figure}

Figure~\ref{fig:attention_viz} illustrates a representative example: the modality-weighted combined attention produces a sharp peak precisely aligned with the ground truth laughter interval, demonstrating successful temporal localization from weak clip-level supervision.

\textbf{Summary.} The ablations confirm that each component (temporal softmax pooling, tanh gating, softmax-based fusion, and projection activation) contributes measurably to both classification and localization performance. The full model represents a carefully tuned balance between temporal precision, modality complementarity, and computational efficiency.

\subsection{Temporal Grounding Enhances Semantic Reasoning}
\label{sec:downstream}

Beyond detection metrics, we evaluate whether precise temporal localization enhances semantic reasoning about laughter. Recent work has highlighted temporal reasoning gaps in multimodal models~\cite{huang2024lita}, particularly for fine-grained social signals. We investigate this through the Video Laugh Reasoning task, where models must explain why audiences laugh.

\textbf{Experimental Results.} Table~\ref{tab:laughter_reasoning} compares GPT-3.5 and GPT-4o performance with and without our temporal tags. For GPT-3.5, providing precise \texttt{<LAUGHTER>} markers yields substantial improvements across all metrics: +28.4\% BLEU-1, +58.7\% BLEU-4, +31.9\% METEOR, +17\% BERTScore, and +227.2\% CIDEr (all statistically significant, $p < 0.001$). Remarkably, GPT-3.5 with temporal tags significantly outperforms the baseline GPT-4o without tags across all metrics, demonstrating that precise temporal grounding can bridge the capability gap between model generations.

For GPT-4o, temporal tags provide positive but smaller improvements (not statistically significant), suggesting a ceiling effect where more capable models may implicitly infer temporal dynamics from semantic context. Nevertheless, even GPT-4o benefits from explicit temporal grounding, particularly on corpus-level metrics (CIDEr: 0.516 vs. 0.563, 9\% improvement).

\textbf{Key Insights.} These results validate that precise temporal localization provides meaningful semantic value. When language models know exactly \textit{when} laughter occurs, they better reason about social dynamics, humor mechanisms, and contextual cues. This finding challenges the trend of relying solely on model advancements for complex reasoning tasks: our lightweight temporal localization enables a less capable model (GPT-3.5) to outperform a state-of-the-art model (GPT-4o) on a reasoning-heavy task. The results demonstrate that task-specific temporal grounding is a viable, and often superior, alternative to relying on general model capabilities, offering improved performance at lower computational cost and latency.

\begin{table}[t]
\centering
\caption{Impact of precise temporal tagging on Video Laugh Reasoning (N=50 SMILE videos). Comparison of GPT-3.5 and GPT-4o with and without temporal markers. Significance: $^{*}p<0.001$ (GPT-3.5 Ours vs. Baseline). $^{\S}$ indicates corpus-level metrics.}
\label{tab:laughter_reasoning}
\resizebox{\columnwidth}{!}{%
\renewcommand{\arraystretch}{1.5}
\begin{tabular}{lccc|cc}
\toprule
\textbf{Model/Method} & \textbf{BLEU-4}$\uparrow$ & \textbf{METEOR}$\uparrow$ & \textbf{BERTScore}$\uparrow$ & \textbf{CIDEr}$^{\S}\uparrow$ & \textbf{ROUGE-L}$^{\S}\uparrow$ \\
\midrule
GPT-3.5 Baseline      & 0.148 & 0.321 & 0.393 & 0.262 & 0.299 \\
GPT-4o Baseline       & 0.164 & 0.365 & 0.400 & 0.516 & 0.325 \\
\midrule
GPT-3.5 + Tags (Ours) & \textbf{0.235}$^{*}$ & \textbf{0.423}$^{*}$ & \textbf{0.459}$^{*}$ & \textbf{0.858} & \textbf{0.388} \\
\bottomrule
\end{tabular}%
}
\end{table}

\section{Conclusion}
\label{sec:conclusion}

We introduced a weakly-supervised framework for precise temporal localization of laughter in multimodal video streams, trained solely from clip-level labels. Across three diverse datasets, our approach achieves strong classification and temporal grounding, consistently outperforming audio-only and audio-visual foundation models. On SMILE, Gemini 3 Flash achieves competitive localization, suggesting that semantically rich content with pronounced audience laughter may partially benefit from general-purpose reasoning, highlighting an interesting boundary case for future investigation.

A core contribution of this work is \textbf{UR-FUNNY-Temporal} and \textbf{SMILE-Temporal} - large-scale benchmarks providing precise laughter annotations for 11,053 videos spanning 78.8 hours. Beyond temporal boundaries, each event is labeled with speaker vs.\ audience origin, modality dominance (acoustic, visual, or both), and intensity level (chuckle vs.\ laughter). Crucially, while our model never uses these annotations during training, we release them in full to enable a broad range of future research directions:

\begin{itemize}
    \item \textbf{Supervised and semi-supervised localization:} our annotations enable training fully-supervised TAL models as strong baselines, and studying how few temporal labels can bootstrap weakly-supervised methods.
    \item \textbf{Modality-aware understanding:} the acoustic/visual/both labels support research on when and why each modality dominates, enabling modality-adaptive models for laughter and other affective events.
    \item \textbf{Speaker vs.\ audience dynamics:} distinguishing who laughs enables studies of social contagion, audience engagement prediction, and speaker-aware affect modeling.
    \item \textbf{Grounded multimodal reasoning:} precise temporal tags can augment transcripts with laughter markers, enabling richer LLM reasoning about humor, sentiment, and conversational dynamics.
\end{itemize}

Beyond detection, we demonstrated that precise temporal tags enhance downstream semantic reasoning, enabling GPT-3.5 to outperform GPT-4o by 227\% on CIDEr, suggesting that temporal grounding is a viable alternative to model scaling for reasoning-heavy tasks.


{\small
\bibliographystyle{ieeenat_fullname}
\bibliography{refs}
}

\end{document}